\documentclass{llncs}

\usepackage{makeidx}
\usepackage[nolist,nohyperlinks]{acronym}
\usepackage[inline]{enumitem}
\usepackage{multirow}
\usepackage{graphicx}
\usepackage{siunitx}
\usepackage{subcaption}
\usepackage{booktabs}
\usepackage[breaklinks=true,bookmarks=false]{hyperref}

\newcommand{\repeatthanks}{\textsuperscript{\thefootnote}}

\begin{document}

\title{Language Identification Using Deep Convolutional Recurrent Neural Networks}
\titlerunning{Language Identification Using CRNNs}
\author{Christian Bartz\thanks{equal contribution}, Tom Herold\repeatthanks, Haojin Yang \and Christoph Meinel}
\authorrunning{Bartz, Herold, Yang, Meinel}

\institute{Hasso Plattner Institute, University of Potsdam \\
	\email{\{christian.bartz, haojin.yang, meinel\}@hpi.de \\ tom.herold@student.hpi.de}}

\frontmatter

\maketitle

\begin{acronym}
	\acro{CNN}{Convolutional Neural Network}
	\acro{RNN}{Recurrent Neural Network}
	\acro{DNN}{Deep Neural Network}
	\acro{BLSTM}{Bidirectional Long-Short Term Memory}
	\acro{LSTM}{Long-Short Term Memory}
	\acro{MFCC}{Mel-Frequency Cepstral Coefficients}
	\acro{SDC}{Shifted Delta Cepstral}
	\acro{UBM}{Unfied Background Model}
	\acro{GMM}{Gaussian Mixture Model}
	\acro{PCA}{Principal Component Analysis}
	\acro{SVM}{Support Vector Machine}
	\acro{PLP}{Perceptual Linear Prediction}
	\acro{MFCC-SDC}{Mel Frequency Cepstral Coefficients with Shifted Delta Coefficients}
	\acro{WAVE}{Waveform Audio File Format}
	\acro{CRNN}{Convolutional Recurrent Neural Network}
	\acro{ASR}{Automatic Speech Recognition}
	\acro{LID}{Language Identification}
\end{acronym}

\begin{abstract}

	\ac{LID} systems are used to classify the spoken language from a given audio sample and are typically the first step for many spoken language processing tasks, such as \ac{ASR} systems.
	Without automatic language detection, speech utterances cannot be parsed correctly and grammar rules cannot be applied, causing subsequent speech recognition steps to fail.
	We propose a \ac{LID} system that solves the problem in the image domain, rather than the audio domain.
	We use a hybrid \ac{CRNN} that operates on spectrogram images of the provided audio snippets.
	In extensive experiments we show, that our model is applicable to a range of noisy scenarios and can easily be extended to previously unknown languages, while maintaining its classification accuracy.
	We release our code and a large scale training set for \ac{LID} systems to the community.
\end{abstract}

\section{Introduction}

Intelligent assistants like Siri\footnote{https://www.apple.com/ios/siri/} or the Google Assistant\footnote{https://assistant.google.com/} rely on \ac{ASR}.
Current \ac{ASR} systems require users to manually specify the system's correct input language to work properly. However, as a sensible pre-processing step we can infer the spoken language using an automatic \ac{LID} system.
Traditional \ac{LID} systems utilize domain-specific expert knowledge in the field of audio signal processing for extracting hand-crafted features from the audio samples.
Lately, deep learning and artificial neural networks have become the state-of-the-art for many pattern recognition problems.
\acp{DNN} have become the best performing method for a range of computer vision tasks, such as image classification~\cite{Simonyan2015Very,Szegedy2016Rethinking}, or object detection and recognition~\cite{Redmon2016You,Ren2015Faster}.

In this paper, we address the problem of language identification from a computer vision perspective.
We extract the target language of a given audio sample by utilizing a hybrid network constructed of a \ac{CNN} combined with an \ac{RNN}.
Our contributions can be summarized as follows:
\begin{enumerate*}[label={(\arabic*)}]
	\item we propose a hybrid \ac{CRNN}, combining the descriptive powers of \acp{CNN} with the ability to capture temporal features of \acp{RNN}.
	\item We perform extensive experiments with our proposed network and show its applicability to a range of scenarios and its extensibility to new languages.
	\item We release our code and a large scale training set for \ac{LID} systems to the community\footnote{https://github.com/HPI-DeepLearning/crnn-lid}.
\end{enumerate*}

The paper is structured in the following way:
In \autoref{sec:related_work} we introduce related work in the field of \ac{LID} systems.
We showcase our system in \autoref{sec:proposed_system} and evaluate it on extensive experiments in \autoref{sec:experiments}.
We conclude our work in \autoref{sec:conclusion}.



\section{Related Work}
\label{sec:related_work}

Traditional language identification systems are based on identity vector systems for spoken-language processing tasks~\cite{Dehak2011FrontEnd,Martinez2011Language,Plchot2016Bat,Zazo2016Language}.
In the recent years systems using feature extractors solely based on neural networks, especially \ac{LSTM} networks, became more popular \cite{Gelly2016DivideAndConquer,Diez2015EndToEnd,Zazo2016Language}.
These neural networks based systems are better suited to \ac{LID} tasks, since they are both simpler in design and provide a higher accuracy than traditional approaches.

\subsection{Identity Vector Systems}

Identity vector systems (i-vectors) have been introduced by Dehak et al.~\cite{Dehak2011FrontEnd} for the purpose of speaker verfication tasks.
i-vectors are a special form of joined low-dimensional representations of speaker and channel factors, as found in earlier joint factor analysis supervectors.

Identity vectors are employed as a data representation in many systems and are fed as inputs to a classifier.
Dehak et al.~\cite{Dehak2011FrontEnd} used \acp{SVM} with cosine kernels.
Other researchers used logistic regression~\cite{Martinez2011Language} or neural networks with three to four layers~\cite{Dominguez2015FrameByFrame,Plchot2016Bat}.
Gelly et al.~\cite{Gelly2016Language} proposed a complex system consisting of phonotactic components~\cite{Zissman1996Comparison}, identity vectors, a lexical system, and \ac{BLSTM} networks for language identification.
The extensive feature engineering with i-vectors results in very complex systems, with an increasing number of computational steps in their pipeline.

\subsection{Neural Network Approaches}

Approaches solely based on applying neural networks on input features like \ac{MFCC} show that they reach state-of-the-art results, while being less complex.

Current research on language identification systems using \acp{DNN} mainly focuses on using different forms of \acp{LSTM}, working on input sequences of transformed audio data.
Zazo et al.~\cite{Zazo2016Language} use \ac{MFCC-SDC} features as input to their unidirectional \ac{LSTM}, which is directly connected to a softmax classifier. The last prediction of the softmax classifier contains the predicted language.
Gelly et al.~\cite{Gelly2016DivideAndConquer} use a \ac{BLSTM} network to capture language information from the input (audio converted to \ac{PLP} coefficients and their first and second order derivatives) in forward and backward direction. The resulting sequence features are fused together and used to classify the language of the input samples.
Both approaches only consider sequences of features as input to their networks.

Lozano-Diez et al.~\cite{Diez2015EndToEnd} perform language identification with the help of \acp{CNN}.
The authors transform the input data to an image containing \ac{MFCC-SDC} features.
The x-axis of that image represents the time-domain and the y-axis describes the individual frequency bins.
Besides plain classification of the input languages with the \ac{CNN}, they also use the \ac{CNN} as feature extractor for identity vectors.
The authors achieve better performance when combining both the \ac{CNN} features and identity vectors.

Our research differs from the mentioned works in the following way:
\begin{enumerate*}[label={(\arabic*)}]
	\item We utilize a strong convolutional feature extractor based on the VGG~\cite{Simonyan2015Very} or Inception-v3~\cite{Szegedy2016Rethinking} architecture.
	\item We use the extracted convolutional features as input to a \ac{BLSTM} and generate our predictions solely based on a deep model.
\end{enumerate*}

\section{Proposed System}
\label{sec:proposed_system}

In our work, we utilize the power of \acp{CNN} to capture spatial information, and the power of \acp{RNN} to capture information through a sequence of time steps for identifying the language from a given audio snippet.
We developed a \ac{DNN} based on a sequence recognition network presented by Shi et al.~\cite{Shi2016EndToEnd}.
In this section, we present the datasets we used for training the network, the audio representation used for training our models, and the structure of our proposed network in detail.

\subsection{Datasets}

Since there are no large-scale, freely available datasets for \ac{LID} tasks (datasets such as the NIST Language Recognition Evaluation are only available behind a paywall), we resorted to creating our own datasets for our experiments.
We collected our datasets from two different sources:
\begin{enumerate*}[label={(\arabic*)}]
	\item we processed speeches, press conferences and statements from the European Parliament, and
	\item we sourced data from news broadcast channels hosted on YouTube.
\end{enumerate*}
We chose to collect data for 6 different languages, while making sure that we include languages with similar phonetics.
Following this idea, we collected data for two Germanic languages (English and German), two Romance languages (French and Spanish), Russian, and Mandarin Chinese.

\paragraph{EU Speech Repository}

The EU Speech Repository\footnote{\url{https://webgate.ec.europa.eu/sr/}} is a collection of video resources for interpretation students.
This dataset is provided for free and consists of debates of the European Parliament, as well as press conferences, interviews, and dedicated training materials from EU interpreters.
Each audio clip is recorded in the speaker's native language and features only one speaker.
The dataset consists of many different female and male speakers.
From this dataset we collected 131 hours of speech data in four languages: English, German, French and Spanish.

\paragraph{YouTube News Collection}

We chose to use news broadcasts as a second data source to obtain audio snippets of similar quality to the EU Speech Repository (different speakers, mostly one speaker at a time and a single defined language).
We gathered all data from YouTube channels such as the official BBC News\footnote{\url{https://www.youtube.com/user/bbcnews}} YouTube channel.

The obtained audio data has many desired properties.
The quality of the audio recordings is very high and hundreds of hours are available online.
News programs often feature guests or remote correspondents resulting in a good mix of different speakers.
Further, news programs feature noise one would expect from a real-world situation: music jingles, nonspeech audio from video clips and transitions between reports.
All in all, we were able to gather 1508 hours of audio data for this dataset.

\subsection{Audio Representation}
\label{subsec:audio_representation}

To make our gathered data compatible with our \ac{LID} system, we need to do some preprocessing.
As a first step, we encode all audio files in the uncompressed, lossless WAVE format, as this format allows for future manipulations without any deterioration in signal quality.
In order to treat our audio snippets as images, we need to transfer the data into the image domain.
We convert our audio data to spectrogram representations for training our models.
The spectrograms are discretized using a Hann~\cite{Blackman1958Measurement} window and 129 frequency bins along the frequency axis (y-axis).
As most phonemes in the English language do not exceed 3 kHz in conversational speech, we only included frequencies of up to 5kHz in the spectrograms.
The time axis (x-axis) is rendered at 50 pixels per second.
We split each audio sequence into nonoverlapping ten-second segments and discard all segments shorter than ten seconds, as we did not want to introduce padding, which might resemble unnatural pauses or silence.
The resulting images are saved as grayscale, lossless $500 \times 129$ PNG files, where frequency intensities are mapped to an eight-bit grayscale range.

\subsection{Architecture}
\label{subsec:architecure}

\begin{figure}[t]
	\includegraphics[width=\textwidth]{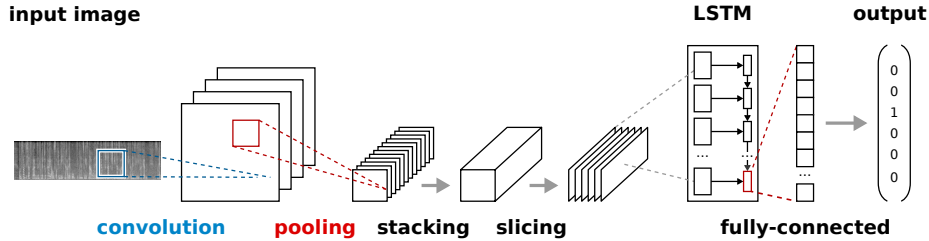}
	\caption{
		Our proposed CRNN network architecture consists of two parts.
		A CNN extracts local visual features from our input images.
		The output of the final convolutional layer $t \times 1 \times c$ is sliced along the time axis into $t$ time steps.
		Each time step represents the extracted frequency features, used as input to the LSTM.
		The final LSTM output is fed into a fully-connected layer for classification.}
	\label{fig:network_structure}
\end{figure}

For our network architecture, we followed the overall structure of the network proposed by Shi et al.~\cite{Shi2016EndToEnd} in their work on scene text recognition.
This network architecture consists of two parts.
The first part is a convolutional feature extractor that takes a spectrogam image representation of the audio file as input (see section~\ref{subsec:audio_representation}).
This feature extractor convolves the input image in several steps and produces a feature map with a height of one.
This feature map is sliced along the x-axis and each slice is used as a time step for the subsequent \ac{BLSTM} network.
The design of the convolutional feature extractor is based on the well known VGG architecture~\cite{Simonyan2015Very}.
Our network uses five convolutional layers, where each layer is followed by the ReLU activation function~\cite{Nair2010Rectified}, BatchNormalization~\cite{Ioffe2015Batcha} and $2 \times 2$ max pooling with a stride of 2.
The kernel sizes and number of filters for each convolutional layer are $(7 \times 7, 16), (5 \times 5, 32), (3 \times 3, 64), (3 \times 3, 128), (3 \times 3, 256)$, respectively.
The \ac{BLSTM} consists of two single \acp{LSTM} with 256 outputs units each.
We concatenate both outputs to a vector of 512 dimensions and feed this into a fully-connected layer with 4/6 output units serving as the classifier.
\autoref{fig:network_structure} provides a schematic overview of the network architecture.

\section{Experiments}
\label{sec:experiments}

Using our gathered dataset and the network architecture introduced in \autoref{sec:proposed_system}, we conducted several experiments, to assess the performance of our proposed network architecture on several kinds of input data.
While performing our experiments we had a range of different questions in mind:

\begin{itemize}
	\item Can we increase the classification accuracy with the help of a network, that combines a \ac{CNN} with a \ac{LSTM}, compared to a \ac{CNN}-only approach?
	\item Is the network able to reliably discriminate between languages?
	\item Is the network robust against different forms of noise in the input data?
	\item Can the network easily be extended to handle other languages as well?
\end{itemize}
First, we shortly introduce our experimental environment and the metrics used.
Then, we show our results on the EU Speech Repository and YouTube News dataset.
Following this, we show the results of our experiments on noise robustness.
We conclude this section with a discussion about the extensibility of our model to new languages.

\subsection{Environment}

We implemented our proposed model using Keras~\cite{Chollet2017Keras} with the Tensorflow~\cite{Abadi2016Tensorflow} backend.
We splitted the datasets into a training (\SI{70}{\percent}), a validation (\SI{20}{\percent}) and a testing set (\SI{10}{\percent}), and all files were distributed equally between the languages.
The European Speech dataset yields a total of about \num{19000} training images, which amounts to roughly 53 hours of speech audio.
The YouTube News dataset yields a total of about \num{194000} training images, or 540 hours of speech audio.
For training our networks, we used the Adam~\cite{Kingma2015Adam} optimizer and resorted to using stochastic gradient descent during fine-tuning.
We observed the following metrics: accuracy, recall, precision and F1 score.
We indicate the used networks in the following way:
\begin{enumerate*}[label={(\arabic*)}]
	\item \ac{CNN} - A network only consisting of the proposed convolutional feature extractor without the recurrent part.
	\item \ac{CRNN} - The proposed hybrid \ac{CRNN} model from section~\ref{subsec:architecure}.
\end{enumerate*}

\subsection{EU Speech Dataset}

\begin{figure}[t]
	\begin{minipage}{0.5\linewidth}
		\centering
		\includegraphics[width=\textwidth]{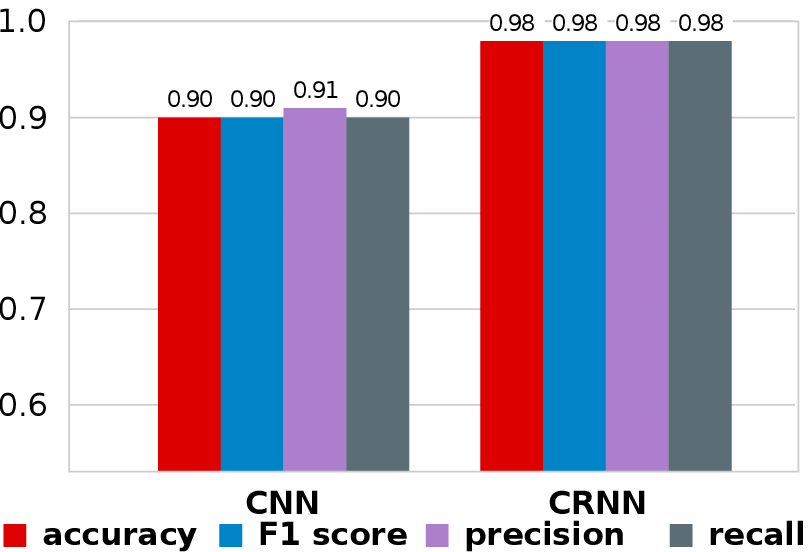}
		\caption{Performance of the proposed network architectures on the EU Speech Repository dataset}
		\label{fig:eu_speech_experiment}
	\end{minipage}\hspace{0.5cm}
	\begin{minipage}{0.5\linewidth}
		\centering
		\includegraphics[width=0.7\textwidth]{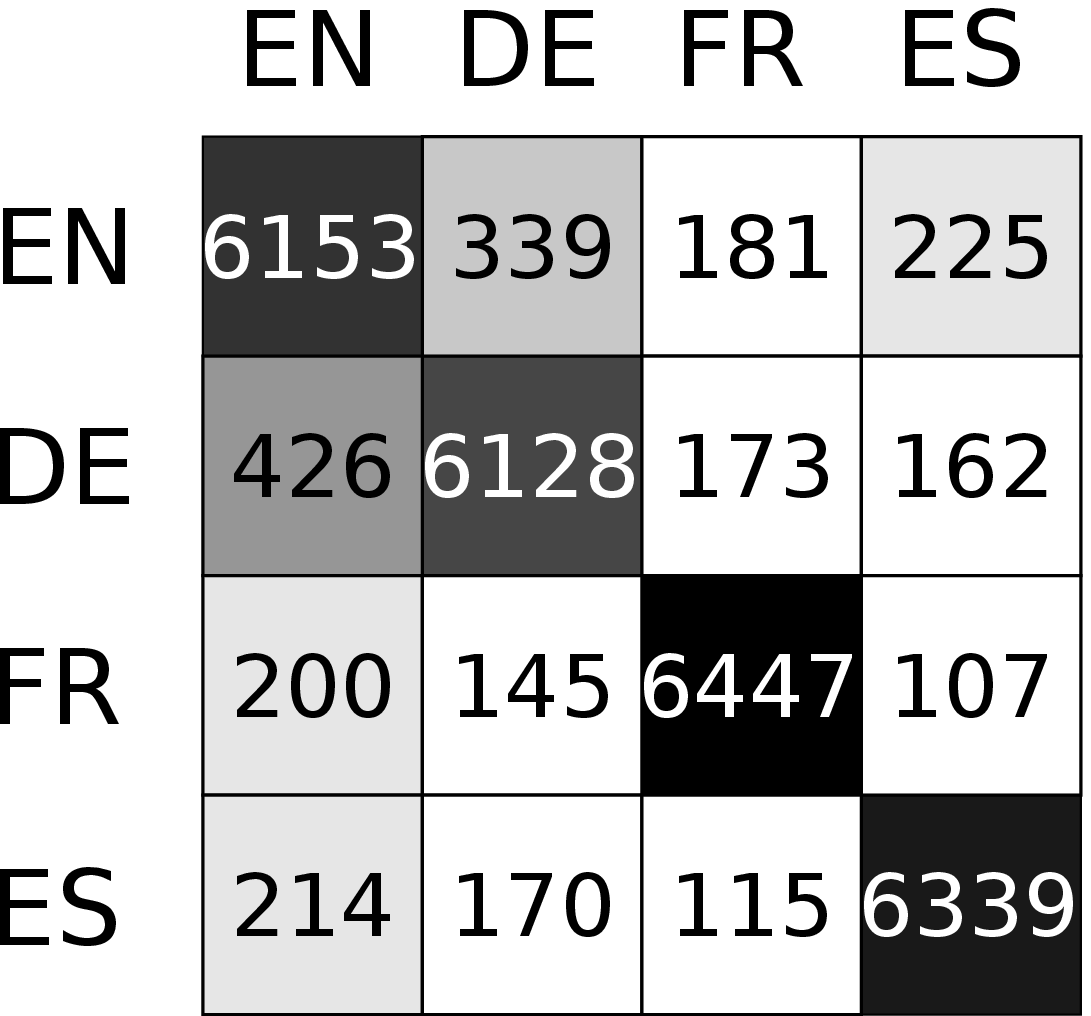}
		\caption{Confusion matrix for our best performing \ac{CRNN}, trained on the YouTube News dataset.}
		\label{fig:confusion_matrix}
	\end{minipage}
\end{figure}

In order to verify our idea of applying \acp{CNN} for classifying image representations of audio data, we established a baseline with the smaller EU Speech Repository dataset.
\autoref{fig:eu_speech_experiment} shows the results for the two network architectures (CNN and CRNN).
As can be seen, the \ac{CRNN} architecture outperforms the plain network without the recurrent part significantly.
This proves our assumption that combining a recurrent network with a convolutional feature extractor increases the accuracy for spoken language identification.

\begin{figure}[t]
	\centering
	\includegraphics[width=\linewidth]{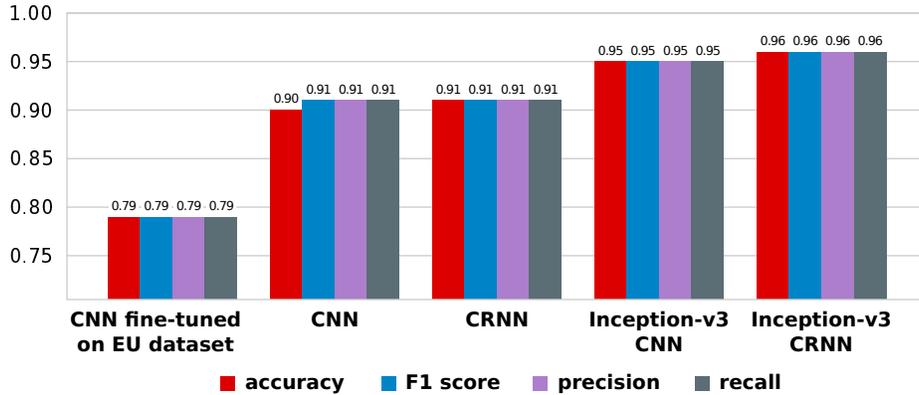}
	\caption{Performance of the trained networks on the YouTube News dataset}
	\label{fig:youtube_experiments}
\end{figure}

\subsection{YouTube News}
\label{subsec:youtube_news}

After achieving these promising results, we trained our networks on the considerably larger YouTube News dataset.
We used only the same four languages that are available in both datasets in order to have comparable  results between these.
First, we took the pre-trained \ac{CNN} from our experiments on the EU Speech Repository dataset and fine-tuned this model with the data from the YouTube dataset.
However, with an accuracy of \SI{79}{\percent}, our \ac{CNN} did not perform as accurately as expected.
We argue that this is the case, because the EU dataset does not feature such a diverse range of situations, as the YouTube News dataset.
Hence, we were unable to take advantage of the already trained convolutional features and the network did not converge on the new dataset.

We trained the \ac{CNN} again, this time from scratch, and were able to achieve an accuracy of \SI{90}{\percent} on this dataset.
Our \ac{CRNN}, using the trained \ac{CNN} as feature extractor, is only able to increase the accuracy to \SI{91}{\percent}.
We argue that the \ac{CNN} already learns to capture some time-related information from the input images.

As a next step, we evaluate how our proposed model architecture compares to a model with a stronger convolutional feature extractor.
Therefore, we trained a model using Google's Inception-v3 layout~\cite{Szegedy2016Rethinking}.
This network architecture is considerably deeper than our proposed \ac{CRNN} architecture and hence should be able to extract more general features.
With Inception-v3 we were able to achieve an accuracy of \SI{95}{\percent} and \SI{96}{\percent} using the \ac{CNN} and \ac{CRNN}, respectively.
These results show that the deeper model is able to catch more general features on our large YouTube dataset. However, this increase in accuracy comes with an increase of computational cost, as the Inception-v3 model uses six times more parameters, than our initially proposed \ac{CNN}.

\autoref{fig:confusion_matrix} shows the confusion matrix when evaluating language family pairs on our best-performing \ac{CRNN}.
Spanish and French separate very well with hardly any wrong classifications.
German and English are more likely to be confused, but English has a slighty stronger bias towards French.
All in all, the learned representations of the model are quite distinctive for each language.

\subsection{Noise Robustness}

In our next series of experiments, we added three different forms of noise to our test data.
We evaluated our already trained models to judge their robustness to these changed conditions.
First, we mixed the audio signal with randomly generated white noise, which has a strong audible presence, but still retains the identifiability of the language.
Second, we added a periodic crackling noise, emulating analog telephony or a bad voice chat connection.
For the last experiment, we added background music from different genres to our samples.

We performed all experiments using our initially proposed \ac{CRNN} and the Inception-v3 \ac{CRNN} architectures.
As expected we observed a decrease in accuracy and F1 score for both models, accross all experiments.
The decrease in accuracy and F1 score is significantly higher for the initially proposed \ac{CRNN} model, but relatively small for the Inception-v3 \ac{CRNN}.
We argue that this is the case because the Inception-v3 \ac{CRNN}, with its deeper and more complex structure, is able to capture the frequency features in a more robust manner.
\autoref{tab:youtube_news_noise_results} shows the results of our experiments with additional noise on the YouTube dataset.

\begin{table}[t]
	\centering
	\begin{tabular}[t]{l c c c c}
		\toprule
		Dataset & \multicolumn{2}{c}{CRNN} & \multicolumn{2}{c}{Inception-v3 CRNN} \\
		 & Accuracy & F1 Score & Accuracy & F1 Score \\
		\midrule
		No Noise & 0.91 & 0.91 & 0.96 & 0.96 \\
		White Noise & 0.63 & 0.63 & 0.91 & 0.91 \\
		Crackling Noise & 0.82 & 0.83 & 0.93 & 0.93 \\
		Background Music & 0.70 & 0.70 & 0.89 & 0.89 \\
		\bottomrule
	\end{tabular}
	\caption{Accuracy and F1 Score measurements for our experiments on the YouTube News dataset with additional noise.}
	\label{tab:youtube_news_noise_results}
\end{table}

\subsection{Extensibility of the network}

\begin{figure}[t]
	\centering
	\includegraphics[width=\linewidth]{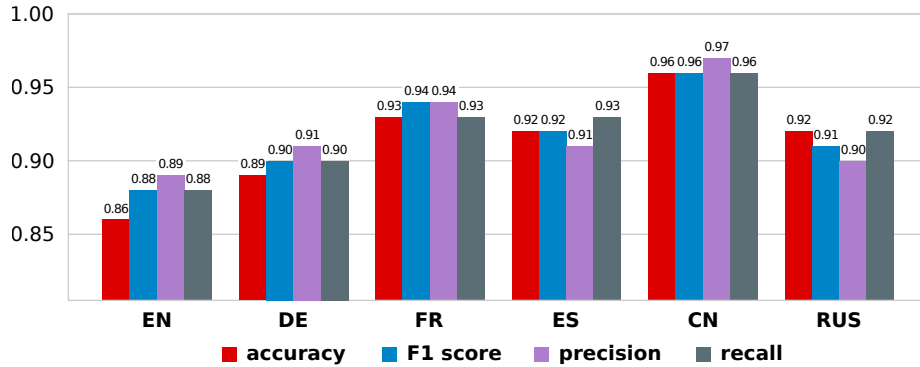}
	\caption{Individual performance measurements for each of our six target languages. Chinese performs best, while English performs worst. Overall, the model performance is consistent with the results of earlier experiments reported in section~\ref{subsec:youtube_news}.}
	\label{fig:youtube_news_extensibility}
\end{figure}

In our last experiment, we evaluated how well our model is able to expand its capabilities to also include new languages, and how adding further languages affects the overall quality of the results produced by the network.
We extended the existing set of four languages by two further languages spoken by millions around the globe: Mandarin Chinese and Russian.
First, we fine-tuned our best performing \ac{CNN} based on our proposed architecture.
The resulting model served as the basis for training the \ac{CRNN} as described earlier.
Applied on the test set, we measure an accuracy of \SI{92}{\percent} and an F1 score of 0.92.
Both measurements match our previous evaluation with four languages on the YouTube News dataset, proving that the proposed \ac{CRNN} architecture can indeed easily be extended to cover more languages. \autoref{fig:youtube_news_extensibility} shows individual performance measures for each language.

We note that Mandarin Chinese outperforms every other language with a top Accuracy of \SI{96}{\percent}, which is likely due to the fact that the sound of Mandarin Chinese is very distinct compared to western languages.
We also find that English is now the worst performing language, which is in part due to a significant number of misclassifications as Russian samples.

In summary, we are content to find that the features learned by our model are universal in nature, as both new languages are rooted within their own language families and feature considerably different intonations.
We believe that our proposed approach to language identification can be successfully applied to a wide variety of languages.

\section{Conclusion}
\label{sec:conclusion}

In this paper we proposed a language identification system, that solves the language identification problem in the image domain, rather than the audio domain.
We proposed a hybrid \ac{CRNN} that consists of a convolutional feature extractor and a \ac{RNN} that combines the extracted features over time.
Using this architecture, we performed several experiments on different datasets to show the wide applicability of our model to various scenarios and its extensibility to new languages.
In order to compensate for the lack of freely available datasets for language identification, we gathered more than 1508 hours of audio data from the EU Speech Repository and YouTube and offer them to the research community.


\bibliographystyle{splncs03}
\bibliography{Remote}

\end{document}